\documentclass[letterpaper, 10 pt, conference]{IEEEconf}

\IEEEoverridecommandlockouts                              

\usepackage{graphics} 
\usepackage{mathtools}
\usepackage{times} 
\usepackage{amsmath} 
\usepackage{amssymb}  
\usepackage{xcolor}
\usepackage[utf8]{inputenc}
\usepackage{comment}
\usepackage{algorithm,algorithmic}
\usepackage{amsthm}
\usepackage{balance}
\usepackage{hyperref}
\DeclareMathAlphabet\mathbfcal{OMS}{cmsy}{b}{n}

\usepackage{cancel}
\theoremstyle{plain}

\newcommand{\q}{\boldsymbol{q}}
\newcommand{\R}{\mathbb{R}}

\newcommand{\x}{\boldsymbol{x}}

\newcommand{\h}{\boldsymbol{h}}
\newcommand{\bxi}{\boldsymbol{\xi}}

\newcommand{\ti}{\text{i}}
\newcommand{\tOA}{\text{OA}}
\newcommand{\tUA}{\text{UA}}
\newcommand{\tFA}{\text{FA}}

\DeclareMathAlphabet{\mathcal}{OMS}{cmsy}{m}{n}

\begin{document}
%
\title{Robust Push Recovery on Bipedal Robots: Leveraging Multi-Domain \\ Hybrid Systems with Reduced-Order Model Predictive Control}

\author{Min Dai and Aaron D. Ames
\thanks{ The authors are with the Department of Mechanical and Civil Engineering, California Institute of Technology, Pasadena, CA 91125 USA.
        {\tt\small \{mdai,ames\}@caltech.edu}.
}}



\maketitle

\begin{abstract}
In this paper, we present a novel control framework to achieve robust push recovery on bipedal robots while locomoting.  The key contribution is the unification of hybrid system models of locomotion with a reduced-order model predictive controller determining: foot placement, step timing, and ankle control.  The proposed reduced-order model is an augmented Linear Inverted Pendulum model with zero moment point coordinates; this is integrated within a model predictive control framework for robust stabilization under external disturbances.  By explicitly leveraging the hybrid dynamics of locomotion, our approach significantly improves stability and robustness across varying walking heights, speeds, step durations, and is effective for both flat-footed and more complex multi-domain heel-to-toe walking patterns. The framework is validated with high-fidelity simulation on Cassie, a 3D underactuated robot, showcasing real-time feasibility and substantially improved stability. The results demonstrate the robustness of the proposed method in dynamic environments.
\end{abstract}


%
\IEEEpeerreviewmaketitle

\section{Introduction}
Achieving robust bipedal locomotion, especially in response to disturbances, is essential for developing robots that operate in complex and unstructured real-world environments.
Yet bipedal locomotion is inherently challenging due to its underlying hybrid dynamics---nonlinear continuous dynamics during the swing phase, and discrete dynamics associated with foot contact \cite{reher_dynamic_2021-1}.  
Inspired by biomechanical studies of human balance, which identified distinct high-level recovery strategies (ankle, hip, and foot placement) to handle external perturbations \cite{nashner1985organization}, researchers have developed similar strategies for humanoid robots. However, most studies have focused on maintaining balance during standing \cite{schuller_online_2021, li_humanoid_2017, stephens_push_2010}, limiting their applicability to locomotion.

When it comes to robotic walking, traditional methods for bipedal locomotion control, such as Zero Moment Point (ZMP) control, have relied heavily on ankle torque to stabilize movement with predetermined footstep locations \cite{kajita_biped_2003}. Under mild walking speeds and with large feet, the robot can be considered as fully-actuated. Thus, with ankle torque as the primary control input, it is often sufficient to handle minor disturbances through tracking controller \cite{vukobratovic_zero-moment_2004, gao_provably_nodate}. However, as walking speed increases or external perturbations become more pronounced, relying solely on ankle torque becomes insufficient.

To address the limitations of ankle-only control, more advanced gait controllers were developed that incorporate foot placement as a critical control input. For instance, the ZMP preview control methods \cite{diedam_online_2008, shafiee-ashtiani_robust_2017} optimize foot placement within a model predictive control (MPC) framework, improving adaptability in dynamic environments. Originally inspired by Raibert's controller \cite{raibert1986legged} for dynamic hopping, foot placement has since been extensively studied in reduced-order models such as the Linear Inverted Pendulum (LIP) and Spring Loaded Inverted Pendulum (SLIP) models \cite{pratt_velocity-based_2006,gong_angular_2021, xiong_3-d_2022, rezazadeh_control_2020}. These models rely on feedback-based foot placement or touch-down angle planning to stabilize walking. Furthermore, Raibert-style regulators have been used in hybrid zero dynamics based approaches \cite{reher_dynamic_2021-1, grizzle_models_2014} to stabilize bipedal robots experimentally, demonstrating the effectiveness of foot placement for maintaining balance.

While foot placement significantly improves robustness in dynamic walking, particularly in reduced-order models like LIP and SLIP, it is often considered in isolation. More recently, researchers have acknowledged that foot placement alone is insufficient to handle larger perturbations. This has led to increased attention on other control inputs, such as step timing and ankle torque, to achieve more comprehensive stability. The inclusion of ankle torque within reduced-order models like LIP model was explored in \cite{acosta_bipedal_2023, dosunmu-ogunbi_stair_2023}, where highly restrictive bounds on ankle torque are used to prevent toe or heel lift-off, thus restricting its role in stabilizing the robot. Similarly, while step timing adjustments have been observed in human walking as a response to disturbances, this technique has mostly been applied heuristically in control frameworks \cite{ghorbani_footstep_2022}. Some Divergent Component of Motion (DCM)-based and Capture Points based approaches have included step timing with foot placement \cite{griffin_walking_2017,mesesan_online_2021,khadiv_walking_2020, kryczka_online_2015}, but none have comprehensively integrated foot placement, step time, and ankle torque into a unified model for robust locomotion. This creates limitations in terms of the adaptability and robustness of the walking strategy as we will show in the results section.

\begin{figure}[t]
    \centering
    \includegraphics[width = 1 \linewidth]{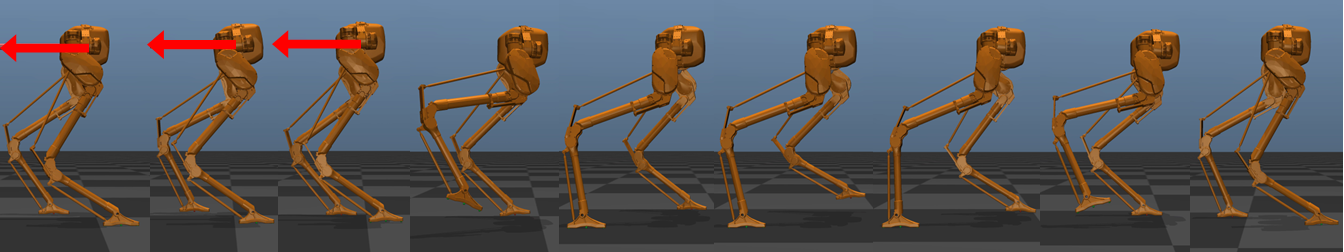}
    \vspace{-5mm}
    \caption{Cassie modifying its step time, foot placement, and using ankle torque to recover from unknown disturbances of -100N for 0.5s. }
    \vspace{-7mm}
    \label{fig::main}
\end{figure}


This paper extends our previous work \cite{dai2024multi}, which introduced a multi-domain walking controller using foot placement stabilized step-to-step dynamics of a custom reduced-order model, by enhancing its application for robust push recovery in flat-footed and multi-domain walking scenarios. In the setting of forward walking, multi-domain walking encompasses the swing foot's heel strike, toe strike, the transition of weight from the new stance foot's heel to toe, and the subsequent heel lift and ankle push-off of the stance foot \cite{ames2011human}. Importantly, multi-domain gaits offer compelling biomechanical advantages \cite{kim_once-per-step_2017, sellaouti_faster_2006} including efficient damping, energy efficiency, faster walking speed, etc. 
Despite these benefits, multi-domain walking is rarely employed in robotic locomotion due to its inherent complexity and fragility induced by the heel lift. To address these challenges, this paper presents an approach that enhances the robustness of multi-domain walking.

In this work, we propose a unified framework for robust bipedal locomotion push recovery, utilizing a variant of the LIP model with an MPC-style planner that generates step-level commands by integrating three critical control inputs: foot placement, step time adjustment, and ankle torque.  To the best of our knowledge, this is the first method that simultaneously employs all three control strategies in an optimization-based framework that significantly increases the push recovery envelope for both flat-footed and multi-domain walking scenarios. 
Moreover, our approach leads to a small-scale nonlinear program that can be solved efficiently, making it suitable for real-time implementation in MPC. We validate the framework through simulations, demonstrating its potential to significantly enhance the stability and robustness of bipedal robots under external perturbations. The collection of the result videos is available in the accompanying video\footnote{\url{https://vimeo.com/1014628632?share=copy}}. 

The remainder of the paper is organized as follows. Section \ref{sec::hybrid_robot} outlines the general hybrid system model for bipedal walking. Section \ref{sec::zlip} introduces the proposed reduced-order model, detailing its continuous and discrete dynamics. These dynamics are then employed to formulate the MPC problem, as described in Section \ref{sec::mpc}. The synthesis of the robot's low-level outputs and the controller setup are discussed in Section \ref{sec::robot_implementation}. Results demonstrating robust push recovery for both flat-footed and multi-domain walking are presented in Section \ref{sec::results}. Finally, conclusions are provided in Section \ref{sec::conclusion}.

\section{Hybrid Dynamics of Bipedal Robots}\label{sec::hybrid_robot}

Bipedal walking can be effectively modeled as a hybrid control system \cite{reher_dynamic_2021-1,grizzle_models_2014}, represented by the tuple $\mathcal{HC} = (\Gamma, \mathcal{D}, \mathcal{S}, \Delta, \mathcal{FG})$, where each component is defined  as:
\begin{itemize}
    \item $\Gamma = (V,E)$ is a directed circle graph with a set of vertices $V = \{ v_\text{i}\}_{\ti \in I}$ and a set of directed edges $E = \{   v_\ti \hspace{-.1cm}\rightarrow \hspace{-.1cm} v_\text{j}   \}_{\ti,\text{j} \in I}$. $I$ is an indexed set of domains.
    \item $\mathcal{D} = \{\mathcal{D}_v\}_{v\in V}$ is a set of domains of admissibility.
    \item $\mathcal{S} = \{ \mathcal{S}_e\}_{e \in E}$ is a set of guards. 
    \item $\Delta = \{ \Delta_e\}_{e \in E}$ is a set of reset maps. 
    \item $\mathcal{FG} = \{f_v, g_v\}_{v\in V}$ is the continuous control system, which is a set of vector fields on the state manifolds.
\end{itemize}

In each domain, the robot's continuous dynamics is governed by the Euler-Lagrange equations, assuming non-slip contact conditions. The equation of motion is given by:
\begin{align}
    &D(\q)\ddot{\q} + H(\q,\dot{\q}) = B \boldsymbol{\tau} + J_\ti(\q)^T \boldsymbol{f}_\ti ,\label{eq::eom} \\ 
    &J_\ti(\q)\ddot{\q} + \dot{J}_\ti(\q,\dot{\q})\dot{\q} = 0, \label{eq::hol} 
\end{align}
where $\boldsymbol{q}\in Q$ is a set of generalized coordinates in the $n$-dimensional configuration space $Q$, $D(\q)\in \R^{n \times n}$, $H(\q,\dot{\q}) \in \R^{n}$, $B\in \R^{n\times m}$ are the inertia matrix, the collection of centrifugal, Coriolis and gravitational forces, and the actuation matrix, respectively. The input torque is denoted by $\boldsymbol{\tau}\in U \subseteq \R^m$. $J_\ti(\q)\in\R^{n\times h_\ti}$ is the domain-specific Jacobian matrix related to contact constraints, and $\boldsymbol{f}_\ti\in\R^{h_\ti}$ represents the corresponding constraint wrench.

Discrete impacts are assumed to be instantaneous and plastic, with solution derivations detailed in \cite{grizzle_models_2014}. Let $\x = [\q^T, \dot{\q}^T]^T \in \mathcal{TQ}$, the  hybrid system's dynamics is given by:
\begin{align}
    \begin{cases} \dot{\x} &= f_v(\x) + g_v(\x) \boldsymbol{\tau}  \quad \x \in \mathcal{D}_v \setminus  \mathcal{S}_e\\
    \x^+ &= \Delta_e( \x^-) \hspace{1.4cm} \x^-\in \mathcal{S}_e  \end{cases},
    \label{eq::HZ}
\end{align}
for all $v \in V $ and corresponding $e \in E$.

\begin{figure}[t]
    \centering
    \vspace{2mm}
    \includegraphics[width = 1 \linewidth]{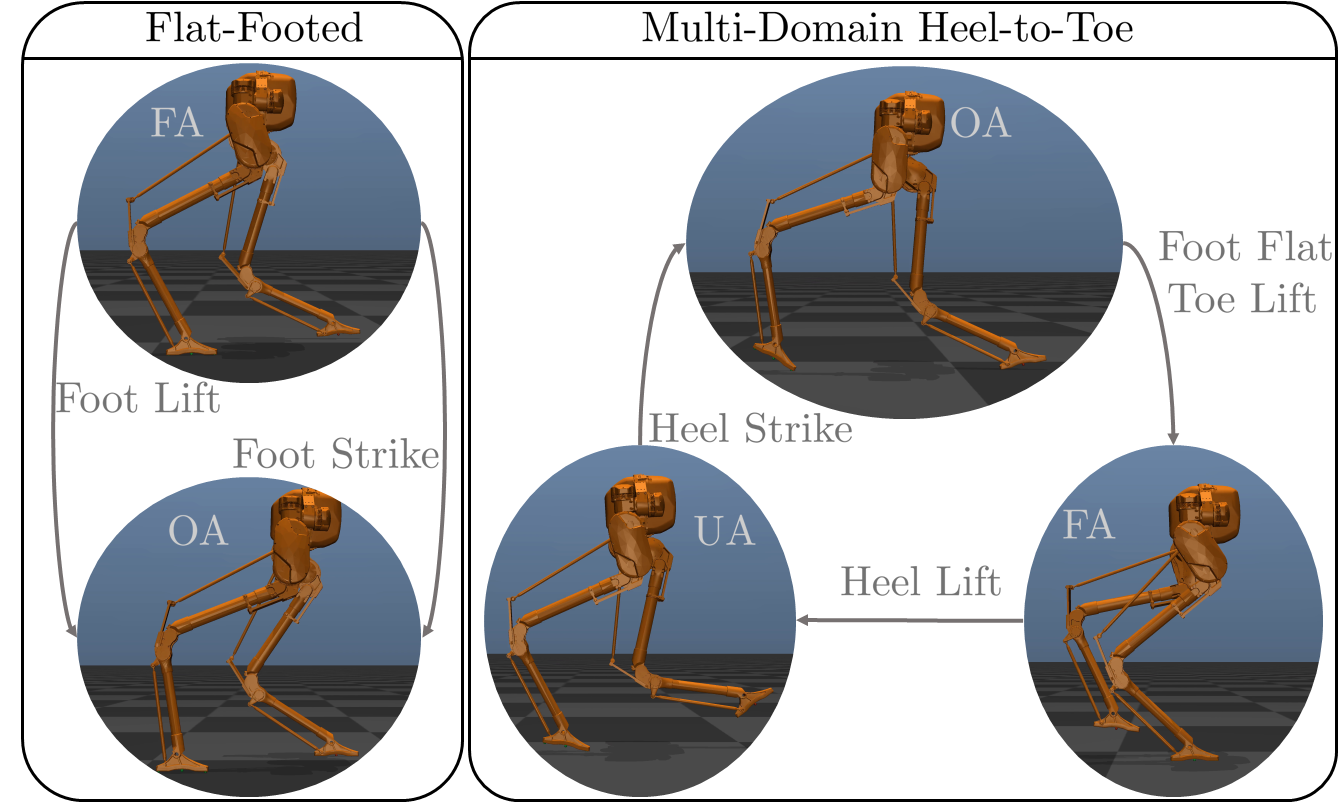}
    \vspace{-5mm}
    \caption{A directed graph showing the hybrid system model used to describe flat-footed and multi-domain bipedal walking.}
    \vspace{-6mm}
    \label{fig::domain_graph}
\end{figure}

Inspired by human heel-to-toe locomotion, we consider a hybrid system model that divides the walking cycle into three distinct domains: fully-actuated (FA), under-actuated (UA), and over-actuated (OA). A graphical representation is given in Fig. \ref{fig::domain_graph}. In the case of heel-to-toe walking, the FA domain represents the single support phase where the stance foot maintains full contact with the ground. The UA domain describes the phase where only the stance toe contacts the ground, and the OA domain corresponds to double support, with both the back leg toe and the front leg heel in contact with the ground. Transitions between these phases are governed by guard conditions based on specific events, such as heel strike or toe-off. For a detailed definition of the hybrid guards, please refer to our earlier work \cite{dai2024multi}.

In this work, we consider both multi-domain walking with underactuated point contact phases as well as regular flat-footed walking. Without loss of generality, flat-footed walking can be considered multi-domain walking with a trivial UA phase and flat-footed OA phase, as shown in Fig. \ref{fig::domain_graph}. This unified model provides a versatile framework that can capture different types of walking behaviors.

\section{Augmented LIP Model with ZMP Control}\label{sec::zlip}

We propose an augmented Linear Inverted Pendulum (LIP) model with ZMP coordinates, referred to as the ZLIP model (Fig. \ref{fig::lip_h2t_defintion}), to account for the effects of a dynamically shifting ZMP during walking. This approach uses ZMP control, which allows for more aggressive use of the ankles while avoiding the risk of unplanned foot tip-overs, a common issue with ankle torque control. This approach also enables explicit constraints for planning heel-lift and toe-lift motions, which is necessary for multi-domian walking. The ZLIP model retains the key features of the LIP model, such as a constant center of mass (CoM) height $z_0$ relative to the stance pivot and the massless legs. The continuous-time dynamics of the ZLIP model is described by:
\begin{figure}[t]
    \centering
    \vspace{2mm}
    \includegraphics[width = 1 \linewidth]{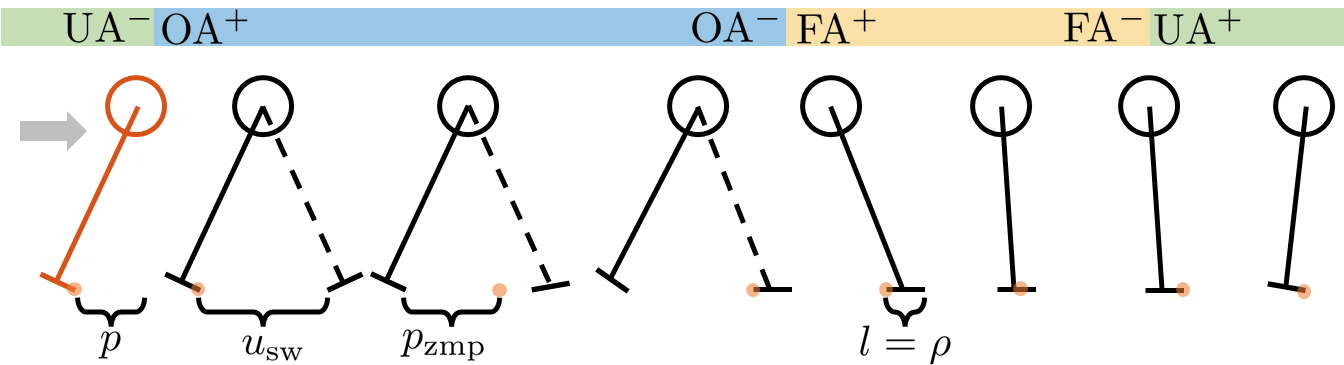}
    \vspace{-5mm}
    \caption{Visualization of the ZLIP model when applied for heel-to-toe walking, highlighting the key phases of the gait cycle. The swing foot location at foot touchdown is denoted as $u_\text{sw}$. During the OA phase, the ZMP (yellow dots) shifts from the back toe to the front heel, while during the FA phase, it moves from heel to toe. The traveled distance in FA phase, $l$, corresponds to the foot curve length $\rho$ for the depicted heel-to-toe walking.}\label{fig::lip_h2t_defintion}
    \vspace{-6mm}
\end{figure}
\begin{align}
\frac{d}{dt}
    \underbrace{\begin{bmatrix}
        p \\ L \\ p_\text{zmp}
    \end{bmatrix}}_{\bxi} =
    \underbrace{\begin{bmatrix}
        0 & \frac{1}{z_0}  & 0\\
        g & 0 &-g \\ 
        0&0&0
    \end{bmatrix} }_{A_\text{ct}}
    \begin{bmatrix}
        p \\ L \\ p_\text{zmp}
    \end{bmatrix}+
    \underbrace{\begin{bmatrix}
        0 \\ 0 \\ 1
    \end{bmatrix}}_{B_\text{ct}} \dot{p}_\text{zmp},
\end{align}
where $p$, $L$, and $p_\text{zmp}$ represent the horizontal CoM position, mass-normalized centroidal angular momentum \cite{orin_centroidal_2013}, and horizontal ZMP position, all defined relative to the stance pivot. In flat-footed walking, the stance pivot can be placed anywhere from heel to toe; here, we select it to be directly under the ankle. For heel-to-toe walking, the stance pivot is positioned at the toe. Since the system dynamics are linear, the end-of-domain states $\bxi_{\ti}^{-}$ has a closed-form solution given by:
\begin{align} \label{eq::MLIP_ct_raw}
    \bxi_\ti^- = \underbrace{e^{A_\text{ct} T_\text{i}}}_{A_\text{i}(T_\text{i})} \bxi_\text{i}^+ + \int_0^{T_\text{i}} e^{A_\text{ct} (T_\text{i}-t)} B_\text{ct} \, \dot{p}_\text{zmp,i}(t) \, dt,
\end{align}
where $\bxi^+$ and $\bxi^-$ indicate the states at the start and end of a domain, respectively, and $T_{\ti}$ represents the duration of the i-th domain. In our previous work, the Multi-domain LIP (MLIP) model was introduced for human-like heel-to-toe walking, where domain durations were predefined and ZMP trajectories were fixed during the FA and OA phases. This resulted in a linear discrete system with passive continuous domain dynamics, leaving foot placement as the only control input. In contrast, the ZLIP model generalizes the MLIP model by treating the ZMP velocity $\dot{p}_\text{zmp,i}(t) = \dot{p}_\text{zmp,i}$ as well as the step duration $T_\ti$ as additional control input:
\begin{align*}
    \int_0^{T_\text{i}} e^{A_\text{ct} (T_\text{i}-t)} B_\text{ct} \, \dot{p}_\text{zmp,i}(t) dt &= \underbrace{\int_0^{T_\text{i}} e^{A_\text{ct} (T_\text{i}-t)} dt B_\text{ct}}_{B_\text{i}(T_\text{i})} \dot{p}_\text{zmp,i}.
\end{align*}

Thus, the actuated continuous domain model is given by:
\begin{align}\label{eq::MLIP_CT}
    \bxi_\ti^- = A_\ti(T_\ti) \bxi_\ti^+ + B_\ti(T_\ti) \dot{p}_\text{zmp,i}.
    \tag{ZLIP-CT}
\end{align}

For impact dynamics, as the model has massless legs, discrete state jumps resulting from rigid body impact do not occur. Instead, transitions between domains in the ZLIP model are purely time-based given virtual legs. The impact equation describes the effects of switching between the stance and swing legs, which occurs at the transition from OA to FA domain. We additionally allow an instantaneous change in ZMP during domain transitions.
Let $B_\text{zmp} = \begin{bmatrix}
    0 & 0 & 1
\end{bmatrix}^T$, $B_u = \begin{bmatrix}
    -1 & 0 & -1
\end{bmatrix}^T$, and $C_l = \begin{bmatrix}
    -l & 0 & -l
\end{bmatrix}^T$, where $l$ is the distance the ZMP travels during the FA phase as shown in Fig. \ref{fig::lip_h2t_defintion}. For heel-to-toe walking, this distance corresponds to the foot curve length, denoted as $\rho$, while for flat-footed walking, $l = 0$ (i.e., no ZMP movement during the FA phase). The impact dynamics equations governing transitions between domains are as follows:
\begin{align} \label{eq::MLIP_DT}
\begin{cases}
   \bxi_\text{OA}^+ &= \bxi_\text{UA}^- + B_\text{zmp} \Delta_\text{zmp}^{\text{UA}\rightarrow\text{OA}}\\
    \bxi_\text{FA}^+ &= \bxi_\text{OA}^- + B_u  u_\text{sw} + C_l + B_\text{zmp} \Delta_\text{zmp}^{\text{OA}\rightarrow\text{FA}}\\
    \bxi_\text{UA}^+ &= \bxi_\text{FA}^-   + B_\text{zmp} \Delta_\text{zmp}^{\text{FA}\rightarrow\text{UA}}
\end{cases}
\tag{ZLIP-DT}
\end{align}
Here, $u_\text{sw} , \Delta_\text{zmp}^{\text{i}\rightarrow\text{i}+1} \in \R$ are control inputs for discrete dynamics of the ZLIP model. 

As shown in Sec. \ref{sec::hybrid_robot}, the structure of the domain graph $\Gamma$ depends on the walking mode. In flat-footed walking, the system simplifies: the UA domain is eliminated ($T_\tUA = 0$), and there is no ZMP travel during the FA phase ($l = 0$). While the presented model in this section focuses on multi-domain walking, the flat-footed variant can be viewed as a special case of the multi-domain model.

\section{MPC Formulation}\label{sec::mpc}
The ZLIP model introduces domain time as a control variable, making the system inherently nonlinear. To address this, we formulate a nonlinear optimization problem that is solved recursively during locomotion. The goal is to optimize the trajectory over a preview horizon of $n$ steps. Leveraging the linear dynamics of the proposed ZLIP model, the nonlinear optimization problem is formulated in a step-to-step manner, eliminating the need for transcription methods like multi-shooting or collocation that typically handle continuous dynamics. This reduction in problem complexity enables efficient real-time execution. Below, we describe the key components of the optimization framework.






\subsection{Decision Variables and Dynamics Constraints}
The state vector $X$ represents the pre-impact states of each domain at every preview step for both the sagittal and coronal planes, i.e. $X_\ti = \begin{bmatrix}
    \bxi_{x,\ti}^T , \bxi_{y,\ti}^T
\end{bmatrix}^T$, as the dynamics in the two planes can be considered as decoupled. To ensure generality across different domains, the same optimization problem structure is used regardless of the current domain. Consequently, the state vector is pre-allocated to account for all possible domains as follows:
\begin{align*}
    X = \begin{Bmatrix}
    X_\text{OA}^0, & 
    X_\text{FA}^0, & 
    X_\text{UA}^0, & 
    \cdots, & 
    X_\text{OA}^n, & 
    X_\text{FA}^n, & 
    X_\text{UA}^n
\end{Bmatrix}.
\end{align*}
Here, \{ \} denotes a vertical stack of vectors for all the terms within the brackets.
The control input vector $U$ includes swing foot positions $U_\text{sw}$, step times $U_T$, and ZMP-related control variables, which will be further explained for different scenarios. Specifically, $U$ is defined as:
\begin{align*}
U = \begin{Bmatrix}
    U_\text{sw}, &
U_T, &U_{\Delta_\text{zmp}}, &U_{\dot{p}_\text{zmp}}, &
    U_\alpha 
\end{Bmatrix},
\end{align*}
where each control inputs $U_{(\cdot)}$ is similarly a vertically stacked sequence: $U_{(\cdot)} = \begin{Bmatrix}
    U_{(\cdot)}^0, & \cdots, & U_{(\cdot)}^n
\end{Bmatrix}$. 

The system’s dynamics are enforced via equality constraints that link the current state and control inputs to the next state, following the sequence:
\begin{align*}
    X_\tOA^k \rightarrow X_\tFA^k \rightarrow X_\tUA^k \rightarrow X_\tOA^{k+1}.
\end{align*}
These constraints capture both the continuous dynamics and the discrete transitions between domains, as described by equations \eqref{eq::MLIP_DT} and \eqref{eq::MLIP_CT}.

For the current domain, the state vector $X_\text{now}$is introduced to capture the robot’s current state, serving as the initial condition for the optimization. This vector only includes horizontal CoM position and mass-normalized angular momentum relative to the stance pivot. We assume the current ZMP position follows the nominal time-based reference trajectory as in \cite{dai2024multi}. However, instantaneous ZMP shifts are allowed via $U_\text{now}$:
\begin{align*}
U_\text{now} = \begin{Bmatrix}
    U_\text{T2imp}, &
    U_{\Delta_\text{zmp},\text{now}}, &
    U_{\dot{p}_\text{zmp},\text{now}},&
    U_{\alpha,\text{now}} 
\end{Bmatrix}.
\end{align*}
Thus, $U_\text{now}$ handles the control inputs for the current step, including the remaining time-to-impact, instantaneous ZMP shifts, and the ZMP rate of change for the current domain. The transition from the current state to the pre-impact states of the current domain is governed by Eq. \eqref{eq::MLIP_CT}, where $T_\ti$ is replaced by the time-to-impact $U_\text{T2imp}$. Consequently, if the current domain is FA or UA, the state $X_\text{OA}^0$ or both $X_\text{OA}^0$ and $X_\text{FA}^0$ are deactivated. $X_\text{now}$ is linked to $X_\text{OA}^0$, $X_\text{FA}^0$ or $X_\text{UA}^0$ depending on the robot's current domain through $U_\text{now}$.

\subsection{Cost Function}
The objective function $J$ is designed to minimize the deviations from a nominal trajectory, as specified by user commands. The nominal reference trajectory is defined by a set of reference parameters, including domain step times, reference velocities in the $x$ and $y$ directions, foot curve length, and CoM height, following the approach in \cite{dai2024multi}. Without loss of generality, we choose the period-1 orbit for the sagittal domain and the period-2 orbit for the coronal plane. The state cost is defined as:
\begin{align*}
    J_X = \sum_{k=0}^{n} \|X_\tUA^k - X^*\|^2_{W_X},
\end{align*}
where $X^* = \begin{bmatrix}
    (\bxi^*)^T & (\bxi_\text{L/R}^*)^T
\end{bmatrix}^T$, with $\bxi^*$ and $\bxi_\text{L/R}^*$ derived as in \cite{dai2024multi} using high-level parameters. The notation $\|X\|_{W}^2\triangleq X^{T}WX$ defines a quadratic cost such that $W$ is a positive-definite diagonal weight matrix. 

Similarly, the cost for control inputs, for the current step and preview steps, is formulated in a quadratic form with weighting matrices $W_\text{sw}$,  $W_T$, $W_{\Delta_\text{zmp}}$ and $W_{\dot{p}_\text{zmp}}$. The reference for the newly introduced decision variable $\Delta_\text{zmp}$ is set to zero. Additionally, for the current step, the step time cost is reformulated as time-to-impact cost:
\begin{align*}
    J_\text{T2imp} = \| U_\text{T2imp} + T_\text{passed} - T_\ti \|_{W_{T_\ti}}^2
\end{align*}
where $T_\text{passed}$ is the time elapsed in the current domain.

\subsection{ZMP constraints} 
\begin{figure}[t]
    \centering
    \vspace{2mm}
    \includegraphics[width = 1 \linewidth]{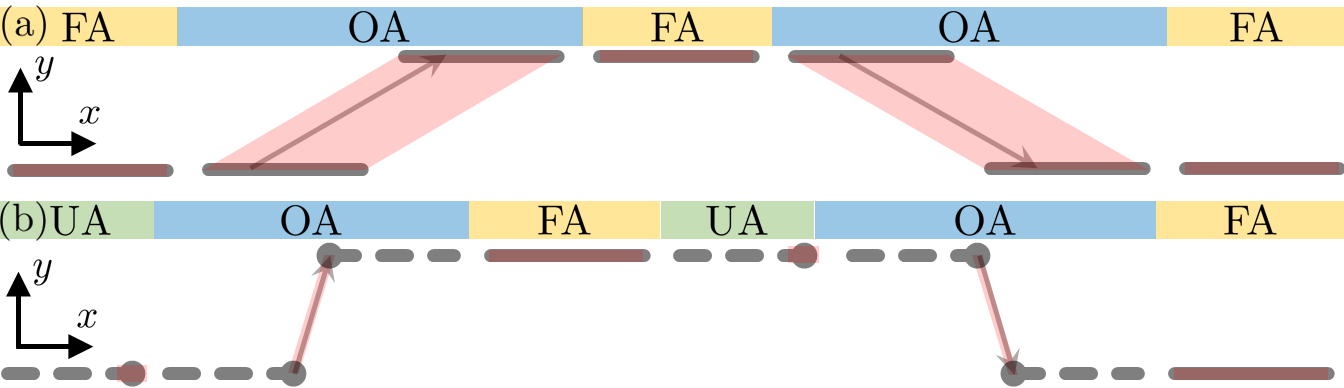}
    \caption{Top-down view of the ZMP constraints as shaded red areas for (a) flat-footed and (b) multi-domain walking for OA (blue), FA (yellow), and UA (green) phases, with foot placement vectors shown in grey.}
    \vspace{-6mm}
    \label{fig::zmp}
\end{figure}
To ensure stability during locomotion, the ZMP must remain within the support polygon throughout each domain. The support polygon is defined as the convex hull of the contact vertices. In general, the ZMP constraints are satisfied if the ZMP position can be expressed as a convex combination of the contact vertices, i.e., $\boldsymbol{p}_\text{zmp} = \sum_{j=1}^{N_v} \lambda_j \boldsymbol{v}_j$ where $\boldsymbol{v}_j$ are the vertices of the support polygon on the ground plane and $\sum_{j=1}^{N_v} \lambda_j = 1$. In this work, the support polygon can take the form of a point, line, or parallelogram. Thus, we simplify the representation as
\begin{align*}
    p_\text{zmp} = \alpha_\text{foot} \begin{bmatrix}
        \rho\\
        0
    \end{bmatrix} + \alpha_\text{step} \begin{bmatrix}
        u_{\text{sw},x} \\
        u_{\text{sw},y}
    \end{bmatrix}, \quad
    \alpha_\text{foot}, \alpha_\text{step} \in [0, 1]
\end{align*}
which can ensure that the ZMP stays within the polygon. For scenarios where the support polygon is a point or line, additional constraints are applied to $\alpha_\text{foot}$ and $ \alpha_\text{step}$ as illustrated in Fig. \ref{fig::zmp}. ZMP constraints are imposed for both pre-impact and post-impact states in each domain to ensure stability throughout the continuous domain, where the post-impact states are derived from Eq. \eqref{eq::MLIP_DT}. The control input $U_\alpha$  is thus defined as:
\begin{align*}
    U_\alpha^k = \begin{bmatrix}
        \alpha_{\text{foot},+}^k &
        \alpha_{\text{step},+}^k &
        \alpha_{\text{foot},-}^k &
        \alpha_{\text{step},-}^k 
    \end{bmatrix}^T.
\end{align*}
If the current domain is OA, the current foot placement $u_\text{sw}$ must be passed to the MPC to enforce ZMP constraints. This unified framework handles both flat-footed and multi-domain walking, with the primary difference being the size and shape of the contact support polygon.

\subsection{Step Time Constraints}
Timing constraints are imposed to regulate the duration of each domain within the walking cycle, ensuring physically realistic movements. These constraints prevent instantaneous foot placements and allow swing foot to reach its desired position within a reasonable time frame. Given the robot's torque limits and requirements to maintain non-slip ground contact conditions, we use $T_\tFA + T_\tUA \geq 0.2$s to ensure sufficient time for the foot to complete its swing. Additionally, the following non-negative constraints are applied: $T_\tFA \geq 0$, $T_\tUA \geq 0$, $T_\tOA \geq 0$.

\subsection{Optimization Setup}
The complete MPC problem is formulated as a nonlinear program (NLP) using the CasADi framework \cite{Andersson2019}. The decision variables $X$, $U$ and $U_\text{now}$, along with the initial condition $X_\text{now}$, high-level reference parameters, the objective function, and all associated constraints, are passed to the IPOPT solver \cite{wachter2006implementation}. The final optimization problem is defined as follows:
\[
\begin{aligned}
& \underset{X, U, U_\text{now}}{\arg\min}
& & J(X,U, U_\text{now}) \\ 
& \text{s.t.}
& & c(X,U, U_\text{now}) = 0 \\
& & & g(X,U,U_\text{now}) \leq 0
\end{aligned}
\tag{ZLIP-MPC} \label{eq::ZLIP-MPC}
\]
where the equality and inequality constraints are derived from the system dynamics and physical limitations discussed earlier. For better convergence under extreme pushes and disturbances, we utilize foot placement, step time, and ZMP control for the first preview step, or equivalently the first three preview domains. For the remaining steps, only foot placement is controlled, as in our previous work \cite{dai2024multi}.

\section{Cassie Implementation}\label{sec::robot_implementation}

\begin{figure}[t]
    \centering
    \vspace{2mm}
    \includegraphics[width = 1 \linewidth]{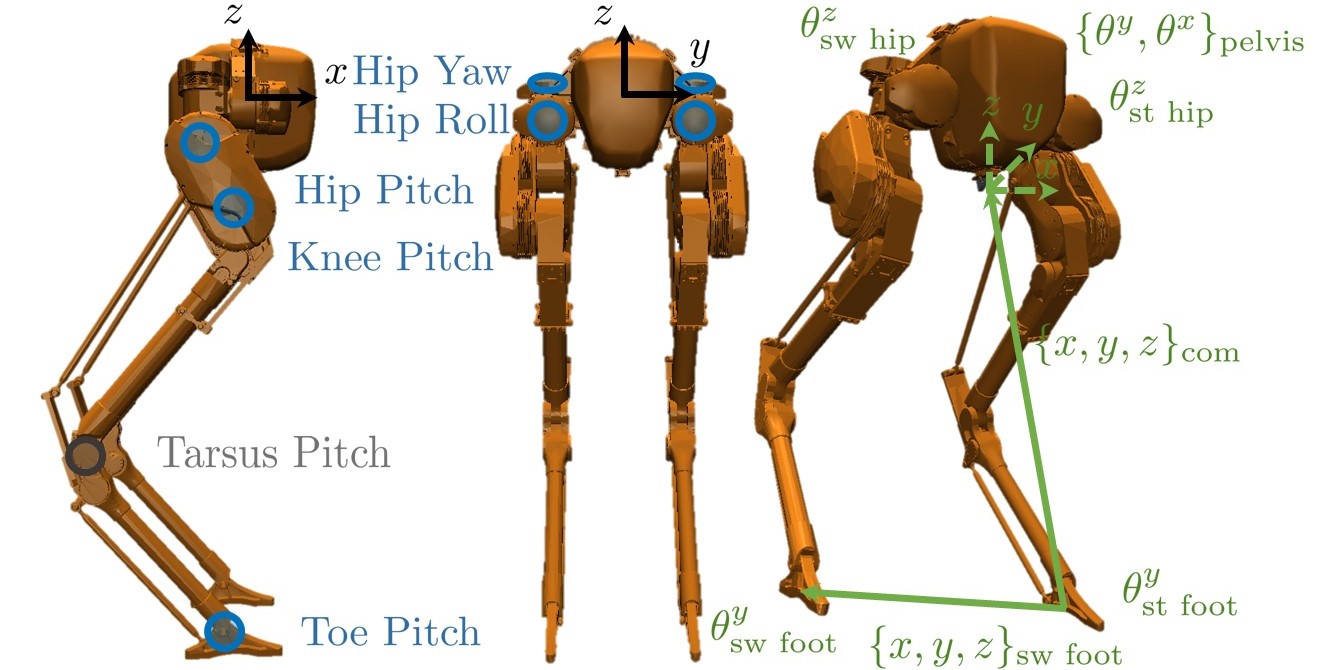}
    \caption{Schematic of Cassie's physical structure and the output definitions used in the control framework.}
    \vspace{-6mm}
    \label{fig::cassie_output}
\end{figure}

We implemented our proposed method on Cassie, a 3D underactuated bipedal robot developed by Agility Robotics \cite{cassie}. As depicted in Fig. \ref{fig::cassie_output}, Cassie has 6 degrees of freedom (DOF) per leg, consisting of five motored joints and one passive tarsus joint. Together with the 6 DOF for the floating-base pelvis, the robot has a total of 18 DOF.

The reduced-order model-based MPC generates commands for foot placement, step duration, and ZMP trajectory in a step-to-step manner. However, to apply these commands to the physical robot in real time, continuous outputs must be constructed, and feedback controllers must be designed accordingly. For detailed output definition for each domain, please refer to \cite{dai2024multi}. This implementation follows the general framework from our previous work. For brevity, the general details of output construction are not repeated here. We focus on two two key modifications: (1) dynamic step time adjustment, and (2) changes to the horizontal CoM reference.

\subsection{Phasing Variable with Changing Step Time}
The desired outputs are constructed using Bézier polynomials parameterized by a phasing variable. When the step time changes, the phasing variable must be adjusted to ensure a smooth, continuous desired output trajectory. To achieve this, we rescale it based on the remaining time in the domain. We use two types of phasing variables for output construction: one for each domain and one for the entire step.

Previously, the phasing variable in each domain is given by $s_\ti = \frac{t_\ti}{T_\ti}$, where $s_\ti(t) \in [0, 1)$ for $t_\ti \in [0, T)$. When a new domain time $T_\ti^j$ is received at $t_\ti^j$ with current phasing variable $s^j$, we need $s_\ti^j(t) \in [s_\ti^j, 1)$ for $t_\ti \in [t^j, T^j)$:
\begin{align}
    s_\ti^j(t) = s_\ti^j + (t_\ti - t_\ti^j) \frac{1-s_\ti^j}{T_\ti^j - t_\ti^j}.\label{eq::s_update}
\end{align}
This formula linearly rescales the remaining time in the domain, ensuring continuity of the phasing variable $s_\ti$. This rescaled phasing variable is primarily used to construct the horizontal CoM trajectories within each domain.

The phasing variable for the entire step, used for outputs like swing foot trajectories and pelvis angles, is defined as:
\begin{align*}
    s = \frac{t}{T_\tFA + T_\tUA} \quad  \quad s \in [0 , 1+\frac{T_\tOA}{T_\tFA + T_\tUA}),
\end{align*}
where $t=0$ marks the start of the FA phase.  Similar to the domain phasing variable, 
$s$ is updated in a way that preserves continuity, but the time spent in already-executed phases remains unchanged. During the FA phase, we update $s$ using Eq. \eqref{eq::s_update}, considering $T_\ti^j = T_\tFA^j + T_\tUA^j$. For the UA domain, $s$ is updated with the same formula but with $T_\ti^j = T_\tFA^\text{passed} + T_\tUA^j$. For the OA phase, $s$ is updated directly using its definition.

\subsection{CoM reference from ZMP trajectory}
Since Cassie lacks foot-mounted force or torque sensors, directly controlling the ZMP is challenging. Instead, we indirectly regulate the ZMP by tracking the CoM states. The reference trajectory for the CoM is defined using a Bézier polynomial, expressed as:
\begin{align}
p_\text{com}^d(s_\ti) \coloneqq b_\text{pcom}(s_\ti) = A(s_\ti) \alpha_\text{pcom},
\end{align}
where $p_\text{com}$ indicates the expression can be used for both $x_\text{com}$ and $y_\text{com}$, and $s_\ti$ is the phase variable in FA or OA domain. $A(s_\ti) \in \R^{1 \times n_b}$  defines the Bézier polynomial of degree $n_b$ and $\alpha_\text{pcom} \in \R^{n_b}$ represents the polynomial coefficients, which are updated during the FA and OA phase subject to the following linear equality constraints:
\begin{align}
    \begin{bmatrix}
        A(s)  \\ A(1) \\ \dot{A}(1,T_\text{i}) 
    \end{bmatrix} \alpha_\text{pcom} = \begin{bmatrix}
        p_\text{com}^a \\ p^{*}_{\text{i}^-} \\ \frac{1}{z_0} L^{*}_{\text{i}^-} 
    \end{bmatrix} ,
\end{align}
where $p^{*}_{\text{i}^-}$ and $L^{*}_{\text{i}^-}$ are desired pre-impact states from NLP solution, such as $X_\tFA^0$ or $X_\tOA^0$. As Cassie has line feet, only $x_\text{com}$ is controlled during FA phase and OA phase for multi-domain walking. Both $x_\text{com}$ and $y_\text{com}$ are controlled in the OA phase for flat-footed walking.

\subsection{Low-level Control}
We employ a task-space quadratic programming (QP) based controller \cite{bouyarmane_quadratic_2019} to ensure the tracking of desired trajectories while respecting the constrained dynamics, physical motor torque limits, and ground contact forces constraints. At each control loop within domain i, we formulate the following QP: 
\begin{align}
   \underset{\boldsymbol{\ddot{q}}, \boldsymbol{\tau}, \boldsymbol{f}_\ti } {\text{min}}  & \quad ||\ddot{\h}^a_\ti(q,\dot{q},\ddot{q}) - \ddot{\h}^d_\ti - \ddot{\h}^t_\ti ||^2_{Q_\ti}, \label{eq::TSC} \tag{TSC-QP} \\
\text{s.t.}  & \quad   \text{Eqs.}~\eqref{eq::eom}, \eqref{eq::hol},  \tag{Dynamics} \\
 & \quad    A_{\text{GRF},\ti} \boldsymbol{f}_\ti  \leq \boldsymbol{b}_{\text{GRF},\ti}, \tag{Contact} \\ 
 & \quad  \boldsymbol{\tau}_{lb} \leq \boldsymbol{\tau} \leq \boldsymbol{\tau}_{ub}.  \tag{Torque Limit}
\end{align}
Here, $Q_\ti$ denotes a weight matrix, $\ddot{\h}^a_\ti$ and $\ddot{\h}^d_\ti$ represent the actual and desired accelerations of the output $\h_\ti$, 
and $ \ddot{\h}^t_\ti = - K_p \h_\ti - K_d \dot{\h}_\ti$ is the target acceleration that enables exponential tracking, with $K_p, K_d$ being the proportional and derivative gains.
The affine contact constraint on $\boldsymbol{f}_\ti$ approximates the contact friction pyramid and ZMP bounds given the contact condition of each domain, while $\boldsymbol{\tau}_{lb}$ and $\boldsymbol{\tau}_{ub}$ are torque limits. Solving the optimization yields the optimal torque $\boldsymbol{\tau}$ for the robot.

\section{Results} \label{sec::results}
\begin{figure}[t]
  \centering
    \vspace{2mm}
    \includegraphics[width=\linewidth]{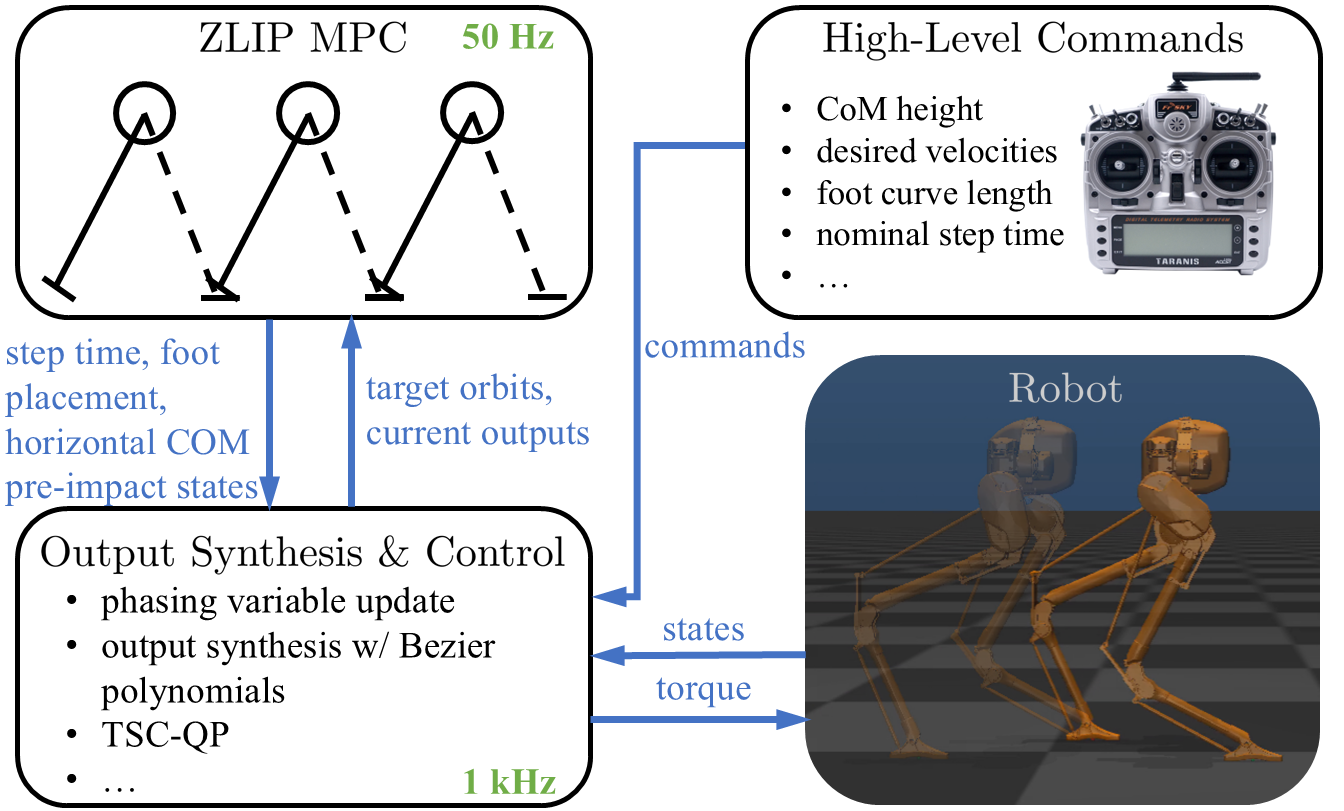}
  \caption{Flowchart summarizing the key components and their interactions in the proposed framework, including the robot and high-level command, the reduced-order model MPC planner, and low-level task-space QP controller for bipedal robot Cassie.}\label{fig::frameworkOVerview}
  \vspace{-6mm}
\end{figure}

We evaluated the proposed approach using our \texttt{C++} framework in the open-source simulator for Cassie \cite{cassie_mujocosim}, which leverages the Mujoco physics engine \cite{todorov_mujoco_2012}. The overall control implementation procedure is summarized in the flowchart in Fig. \ref{fig::frameworkOVerview}. High-level commands, such as CoM height and desired horizontal velocities, are passed to the output synthesis module. This module then processes the commands and parses the desired orbits and other output-related parameters, like time elapsed in the current domain, to the MPC. 
A preview of $n=2$ steps is used for \cite{zaytsev_two_2015} the ZLIP-based MPC planner \ref{eq::ZLIP-MPC}, which is solved at a frequency of $50$Hz. The average MPC solve time was $8$ms on a desktop computer with an Intel® Core™ i7-10700KF CPU @ 3.80GHz and 32GB of RAM, without any specific computation optimization. This framework can be easily extended to handle more preview steps if necessary. The solver is first initialized with reference states and inputs, then warm-started using the previous solution. The output construction and the corresponding low-level controller, as described in \ref{eq::TSC}, are executed at a rate of 1kHz for real-time control. A visual representation of the results can be found in the provided supplementary video.

For all testing scenarios, we assumed a constant CoM height of $z_0 = 0.8$m and a foot length of $\rho = 0.16$m, consistent with Cassie's physical foot dimensions. The push recovery performance was evaluated with unknown disturbances from different directions, without prior knowledge provided to the MPC or the low-level task-space QP controller.


\subsection{Flat-Footed Walking}

\begin{figure}[t]
\centering
    \vspace{2mm}
    \includegraphics[width=\linewidth]{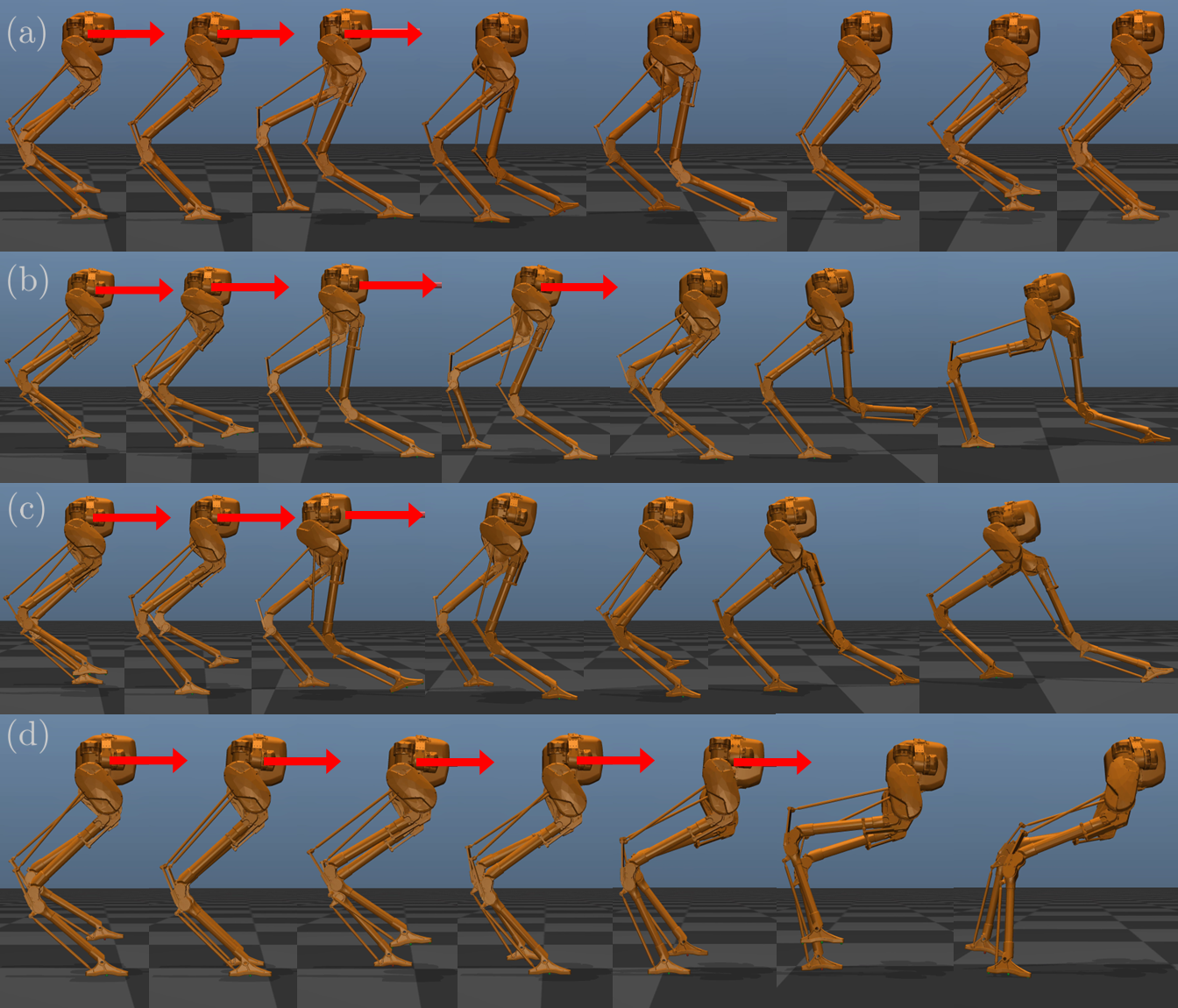}
  \vspace{-4mm}
  \caption{Ablation study of Cassie's response to a 130N sagittal push under different control strategies: (a) proposed method with foot placement, step time, and ZMP control; (b) no ZMP control; (c) no step time control; and (d) no foot placement control.}\label{fig::gaittile_comparision}
  \vspace{-2mm}
\end{figure}

\begin{figure}[t]
  \centering\includegraphics[width=1\linewidth]{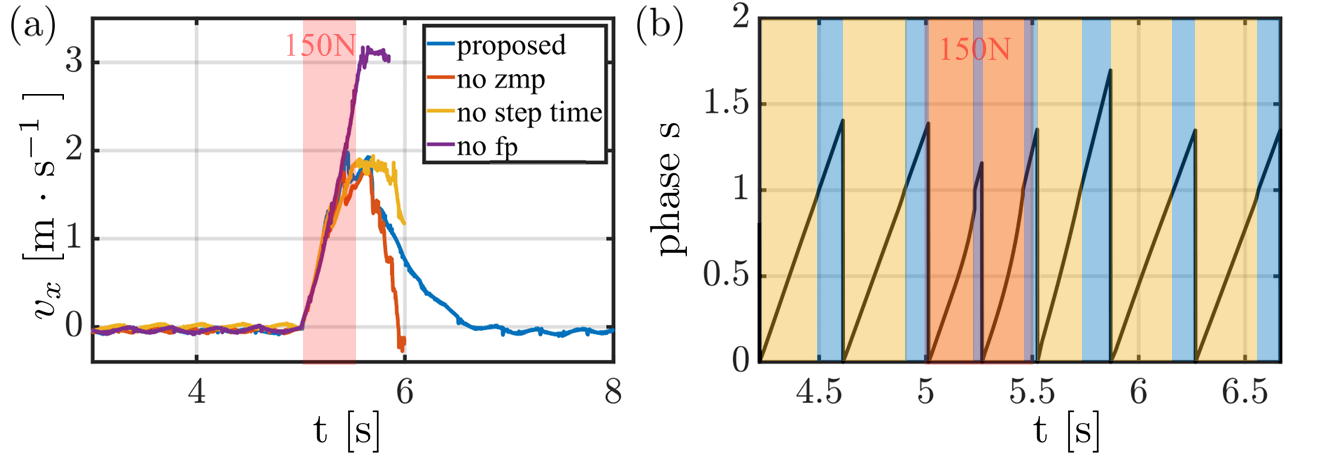}
  \vspace{-6mm}
  \caption{(a) Resultant sagittal CoM velocities and (b) phasing variable trajectory under disturbances. The yellow and blue shaded areas represent FA and OA phases, respectively. }\label{fig::sag_data}
  \vspace{-6mm}
\end{figure}
We first evaluated the algorithm on standard flat-footed walking. In all tests, the step times were set to $T_\tFA = 0.3$s and $T_\tOA = 0.1$s. 
The robot was commanded to walk in place, and a 130N force was applied at the pelvis for 0.5s to test the push recovery performance. 

Fig. \ref{fig::gaittile_comparision} presents a gait tile comparison of Cassie’s response under four different control strategies: the proposed method with foot placement, step time, and ZMP control; a version with no ZMP control; a version with no step time control; and a version with no foot placement control. The ``no ZMP'' version was implemented by forcing $U_\alpha$ to follow nominal assumptions. The ``no step time'' controller fixed the bounds on the control inputs  $U_T$ and $U_\text{T2imp}$. Finally, the ``no foot placement'' controller constrained $U_\text{sw}$ to use a nominal foot placement with a $5$cm relaxation for feasibility.

As shown in Fig. \ref{fig::gaittile_comparision}, only the proposed method can stabilize walking under such extreme disturbances. The foot placement + step time (Fig. \ref{fig::gaittile_comparision}(b)) and the foot placement + ZMP case (Fig. \ref{fig::gaittile_comparision}(c)) was able to partially recover from the disturbance. However, the disturbance was too large that the commanded foot placement for the second step exceeds the robot's kinematic limit. For the case with only step time and ZMP control (Fig. \ref{fig::gaittile_comparision}(d)), although the robot increased stepping frequency and commanded ZMP to be at the tip of the robot, the CoM after disturbance was already outside of the support polygon. Thus, the robot cannot recover without stepping its foot away. 

Fig. \ref{fig::sag_data}(a) shows the resultant CoM velocities for the ablation study. Additionally, the resultant phasing variable from the proposed method is also provided in Fig. \ref{fig::sag_data}(b) with OA phase shaded in blue and FA phase shaded in yellow. When step time control was activated, the MPC reduced the FA and OA phase duration in response to the disturbance, similar to human walking behavior under external perturbations.

\begin{figure}[t]
    \vspace{2mm}
  \centering\includegraphics[width=1\linewidth]{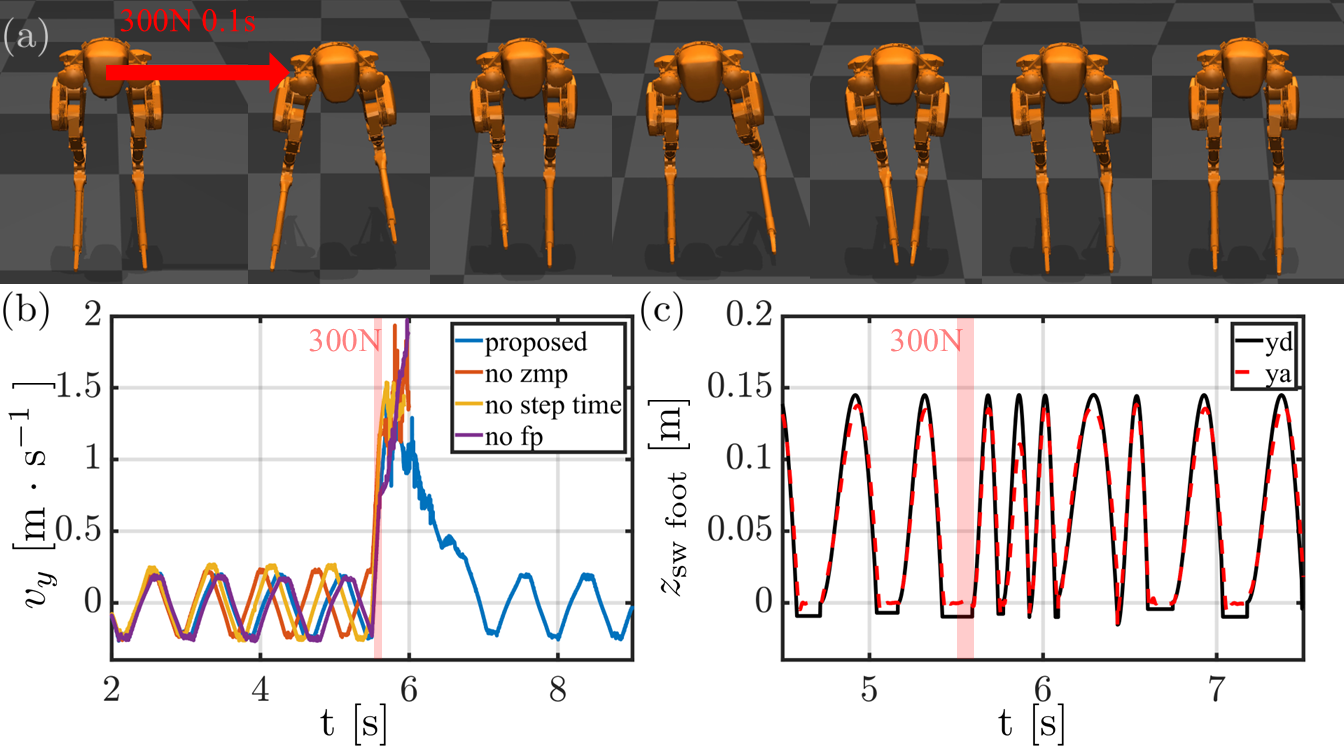}
  \vspace{-6mm}
  \caption{(a) Cassie recovers from a 300N lateral push. (b) Resultant CoM velocities in $y$ plane for different controllers. (c) Resultant swing foot z-trajectory for proposed method during push recovery.}\label{fig::lateral_results}
  \vspace{-7mm}
\end{figure}

We additionally tested Cassie with a 300N lateral push for 0.1s. As shown in Fig. \ref{fig::lateral_results}(b), the proposed method was the only one capable of stabilizing the robot. Despite the push increased the robot’s CoM velocity to nearly 2m/s, the framework successfully stabilized the robot, allowing it to return to walking in place. Fig. \ref{fig::lateral_results}(c) shows the resultant swing foot $z$ trajectory during the push recovery. The reference swing $z$ trajectory is constructed as a Bézier polynomial that reaches a small negative value (0.01m in this case) at the end of the FA phase, ensuring proper foot strike. The swing foot trajectory in OA phase is shown for visualization purposes. As in the OA phase, the swing foot positions are included in the holonomic constraints instead of the output. As shown in the figure, even when the step time was reduced, the proposed rescaling of the phasing variable ensured that the swing foot reached the ground at the commanded time.





\subsection{Multi-Domain Walking}
The proposed methodology can also be applied to human-like multi-domain walking, incorporating the heel-to-toe transition. In all tests, we used $T_\text{SS} = 0.4$s and $T_\text{OA} = 0.1$s. For the ZLIP model in the sagittal plane, we set $T_\tFA = T_\tUA = \frac{T_\text{SS}}{2}$, resulting in a phase distribution of 40\% FA, 40\% UA, and 20\% OA, consistent with human walking data \cite{birch_terminology_2015}. In the lateral plane, the time distribution was adjusted to $T_\tFA = 0$ and $ T_\tUA = T_\text{SS}$ to match Cassie's line feet.

\begin{figure}[t]
  \vspace{2mm}
  \centering\includegraphics[width=\linewidth]{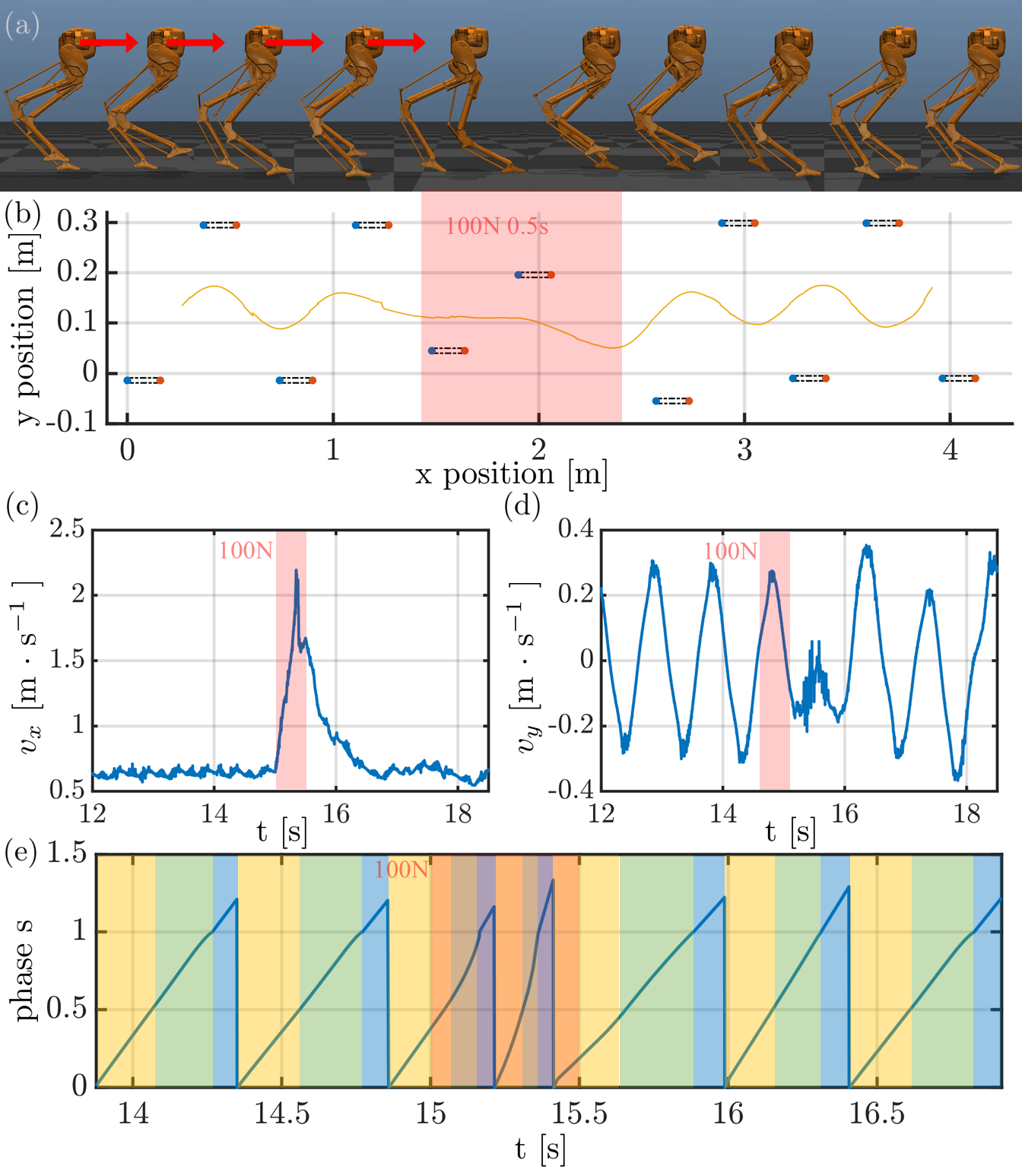}
  \vspace{-7mm}
  \caption{(a) Cassie recovers from a 0.5s, 100N push while performing heel-to-toe walking (b) Top-down view of foot placement and CoM trajectory during disturbance recovery. Blue and red dots indicate heel strike and toe push-off locations, respectively, and the CoM trajectory is shown in yellow. (c) Horizontal velocity profiles under disturbance for sagittal and coronal planes. and (e)  Step-level phasing variable with UA, OA, and FA phases shaded in green, blue, and yellow.}\label{fig::multidomain_result}
  \vspace{-7mm}
\end{figure}

Fig. \ref{fig::multidomain_result}(a) shows the gait sequence during forward walking, where the robot recovered from a 100N push applied for 0.5s. Fig. \ref{fig::multidomain_result}(b) provides a top-down view of the recovery process, illustrating the heel touchdown and ankle push-off locations, along with the resultant CoM trajectory. Note that lateral steps were also adjusted due to the reduced step time, even though the disturbance was in the sagittal direction.

The horizontal velocity profiles for the sagittal and coronal planes are shown in Fig. \ref{fig::multidomain_result}(c) and (d). Multi-domain walking tends to be less robust than flat-footed walking due to the presence of the UA phase. As illustrated in Fig. \ref{fig::multidomain_result}(e), the phasing variable during recovery reveals that the rate of change in sagittal velocity during the UA phase (Fig. \ref{fig::multidomain_result}(c)) was significantly higher than in the FA and OA phases when under disturbances. Similar to flat-footed walking, the step times for the FA, UA, and OA phases were reduced when subjected to disturbances. For additional results, please refer to our complimentary video.


\section{Conclusion and Discussion}\label{sec::conclusion}

We proposed a generalized framework for robust bipedal locomotion, combining foot placement, step timing adjustment, and ankle torque in a computationally efficient MPC-based control strategy, enabling more robust and stable walking behaviors.  The effectiveness of this framework was validated on Cassie, demonstrating significant improvements in push recovery and stability for both flat-footed and multi-domain walking. The results showcase the potential of this approach to enhance robustness under large perturbations.


Future work will focus on extending the method presented in this paper to a full-body humanoid with an upper body and arms. 
The robot used throughout this paper, Cassie, is a lower-body biped without arms and has limited capability in regulating its centroidal angular momentum \cite{orin_centroidal_2013}. The framework presented does not include centroidal momentum regulation, which is typically considered hip strategy \cite{schuller_online_2021, kojio_footstep_2020} in the push-recovery literature. However, the reduced order model presented can be extended to include centroidal momentum regulation by adding $\dot{L}_\text{com}$ as control input in continuous domain with $B_\text{L} = \begin{bmatrix} 0 & 1 & 0 \end{bmatrix}^T$. We believe this extension will enable the application of the reduced-order model MPC to full-body humanoid robots.

\bibliographystyle{IEEEtran}
\balance

\bibliography{references,misc}

\begin{thebibliography}{10}
\providecommand{\url}[1]{#1}
\csname url@samestyle\endcsname
\providecommand{\newblock}{\relax}
\providecommand{\bibinfo}[2]{#2}
\providecommand{\BIBentrySTDinterwordspacing}{\spaceskip=0pt\relax}
\providecommand{\BIBentryALTinterwordstretchfactor}{4}
\providecommand{\BIBentryALTinterwordspacing}{\spaceskip=\fontdimen2\font plus
\BIBentryALTinterwordstretchfactor\fontdimen3\font minus \fontdimen4\font\relax}
\providecommand{\BIBforeignlanguage}[2]{{%
\expandafter\ifx\csname l@#1\endcsname\relax
\typeout{** WARNING: IEEEtran.bst: No hyphenation pattern has been}%
\typeout{** loaded for the language `#1'. Using the pattern for}%
\typeout{** the default language instead.}%
\else
\language=\csname l@#1\endcsname
\fi
#2}}
\providecommand{\BIBdecl}{\relax}
\BIBdecl

\bibitem{reher_dynamic_2021-1}
J.~Reher and A.~D. Ames, ``\BIBforeignlanguage{en}{Dynamic {Walking}: {Toward} {Agile} and {Efficient} {Bipedal} {Robots}},'' \emph{\BIBforeignlanguage{en}{Annual Review of Control, Robotics, and Autonomous Systems}}, vol.~4, no.~1, pp. 535--572, 2021.

\bibitem{nashner1985organization}
L.~M. Nashner and G.~McCollum, ``The organization of human postural movements: a formal basis and experimental synthesis,'' \emph{Behavioral and brain sciences}, vol.~8, no.~1, pp. 135--150, 1985.

\bibitem{schuller_online_2021}
R.~Schuller, G.~Mesesan, J.~Englsberger, J.~Lee, and C.~Ott, ``Online {Centroidal} {Angular} {Momentum} {Reference} {Generation} and {Motion} {Optimization} for {Humanoid} {Push} {Recovery},'' \emph{IEEE Robotics and Automation Letters}, vol.~6, no.~3, pp. 5689--5696, Jul. 2021, conference Name: IEEE Robotics and Automation Letters.

\bibitem{li_humanoid_2017}
Z.~Li, C.~Zhou, Q.~Zhu, and R.~Xiong, ``Humanoid {Balancing} {Behavior} {Featured} by {Underactuated} {Foot} {Motion},'' \emph{IEEE Transactions on Robotics}, vol.~33, no.~2, pp. 298--312, Apr. 2017, conference Name: IEEE Transactions on Robotics.

\bibitem{stephens_push_2010}
B.~J. Stephens and C.~G. Atkeson, ``\BIBforeignlanguage{en}{Push {Recovery} by stepping for humanoid robots with force controlled joints},'' in \emph{\BIBforeignlanguage{en}{2010 10th {IEEE}-{RAS} {International} {Conference} on {Humanoid} {Robots}}}.\hskip 1em plus 0.5em minus 0.4em\relax Nashville, TN, USA: IEEE, Dec. 2010, pp. 52--59.

\bibitem{kajita_biped_2003}
S.~Kajita, F.~Kanehiro, K.~Kaneko, K.~Fujiwara, K.~Harada, K.~Yokoi, and H.~Hirukawa, ``Biped walking pattern generation by using preview control of zero-moment point,'' in \emph{Proceedings of the 2003 {IEEE} {International} {Conference} on {Robotics} \& {Automation}}, vol.~2, Taipei, Taiwan, Sep. 2003.

\bibitem{vukobratovic_zero-moment_2004}
M.~Vukobratović and B.~Borovac, ``Zero-moment point — thirty five years of its life,'' \emph{International Journal of Humanoid Robotics}, vol.~01, no.~01, pp. 157--173, Mar. 2004.

\bibitem{gao_provably_nodate}
Y.~Gao, K.~Barhydt, C.~Niezrecki, and Y.~Gu, ``\BIBforeignlanguage{en}{Provably {Stabilizing} {Global}-{Position} {Tracking} {Control} for {Hybrid} {Models} of {Multi}-{Domain} {Bipedal} {Walking} via {Multiple} {Lyapunov} {Analysis}}.''

\bibitem{diedam_online_2008}
H.~Diedam, D.~Dimitrov, P.-B. Wieber, K.~Mombaur, and M.~Diehl, ``Online walking gait generation with adaptive foot positioning through {Linear} {Model} {Predictive} control,'' in \emph{2008 {IEEE}/{RSJ} {International} {Conference} on {Intelligent} {Robots} and {Systems}}, Sep. 2008, pp. 1121--1126, iSSN: 2153-0866.

\bibitem{shafiee-ashtiani_robust_2017}
M.~Shafiee-Ashtiani, A.~Yousefi-Koma, and M.~Shariat-Panahi, ``Robust bipedal locomotion control based on model predictive control and divergent component of motion,'' in \emph{2017 {IEEE} {International} {Conference} on {Robotics} and {Automation} ({ICRA})}, May 2017, pp. 3505--3510.

\bibitem{raibert1986legged}
M.~H. Raibert, \emph{Legged robots that balance}.\hskip 1em plus 0.5em minus 0.4em\relax MIT press, 1986.

\bibitem{pratt_velocity-based_2006}
J.~Pratt and R.~Tedrake, ``\BIBforeignlanguage{en}{Velocity-{Based} {Stability} {Margins} for {Fast} {Bipedal} {Walking}},'' in \emph{\BIBforeignlanguage{en}{Fast {Motions} in {Biomechanics} and {Robotics}}}.\hskip 1em plus 0.5em minus 0.4em\relax Berlin, Heidelberg: Springer, 2006, vol. 340, pp. 299--324.

\bibitem{gong_angular_2021}
Y.~Gong and J.~Grizzle, ``Angular {Momentum} about the {Contact} {Point} for {Control} of {Bipedal} {Locomotion}: {Validation} in a {LIP}-based {Controller},'' Apr. 2021, arXiv:2008.10763 [cs, eess].

\bibitem{xiong_3-d_2022}
X.~Xiong and A.~Ames, ``3-{D} {Underactuated} {Bipedal} {Walking} via {H}-{LIP} {Based} {Gait} {Synthesis} and {Stepping} {Stabilization},'' \emph{IEEE Transactions on Robotics}, vol.~38, no.~4, pp. 2405--2425, Aug. 2022.

\bibitem{rezazadeh_control_2020}
S.~Rezazadeh and J.~W. Hurst, ``Control of {ATRIAS} in three dimensions: {Walking} as a forced-oscillation problem,'' \emph{The International Journal of Robotics Research}, vol.~39, no.~7, pp. 774--796, Jun. 2020.

\bibitem{grizzle_models_2014}
J.~W. Grizzle, C.~Chevallereau, R.~W. Sinnet, and A.~D. Ames, ``\BIBforeignlanguage{en}{Models, feedback control, and open problems of {3D} bipedal robotic walking},'' \emph{\BIBforeignlanguage{en}{Automatica}}, vol.~50, no.~8, pp. 1955--1988, Aug. 2014.

\bibitem{acosta_bipedal_2023}
B.~Acosta and M.~Posa, ``Bipedal {Walking} on {Constrained} {Footholds} with {MPC} {Footstep} {Control},'' Sep. 2023, arXiv:2309.07993 [cs].

\bibitem{dosunmu-ogunbi_stair_2023}
O.~Dosunmu-Ogunbi, A.~Shrivastava, G.~Gibson, and J.~W. Grizzle, ``Stair {Climbing} using the {Angular} {Momentum} {Linear} {Inverted} {Pendulum} {Model} and {Model} {Predictive} {Control},'' Jul. 2023, arXiv:2307.02448 [cs, eess].

\bibitem{ghorbani_footstep_2022}
E.~Ghorbani, H.~Karimpour, V.~Pasandi, and M.~Keshmiri, ``\BIBforeignlanguage{en}{Footstep {Adjustment} for {Biped} {Push} {Recovery} on {Slippery} {Surfaces}},'' \emph{\BIBforeignlanguage{en}{Multibody System Dynamics}}, vol.~56, no.~3, pp. 189--219, Nov. 2022, arXiv:2111.05203 [cs, eess].

\bibitem{griffin_walking_2017}
R.~J. Griffin, G.~Wiedebach, S.~Bertrand, A.~Leonessa, and J.~Pratt, ``Walking stabilization using step timing and location adjustment on the humanoid robot, {Atlas},'' in \emph{2017 {IEEE}/{RSJ} {International} {Conference} on {Intelligent} {Robots} and {Systems} ({IROS})}, Sep. 2017, pp. 667--673, iSSN: 2153-0866.

\bibitem{mesesan_online_2021}
G.~Mesesan, J.~Englsberger, and C.~Ott, ``\BIBforeignlanguage{en}{Online {DCM} {Trajectory} {Adaptation} for {Push} and {Stumble} {Recovery} during {Humanoid} {Locomotion}},'' in \emph{\BIBforeignlanguage{en}{2021 {IEEE} {International} {Conference} on {Robotics} and {Automation} ({ICRA})}}.\hskip 1em plus 0.5em minus 0.4em\relax Xi'an, China: IEEE, May 2021, pp. 12\,780--12\,786.

\bibitem{khadiv_walking_2020}
M.~Khadiv, A.~Herzog, S.~A.~A. Moosavian, and L.~Righetti, ``Walking {Control} {Based} on {Step} {Timing} {Adaptation},'' \emph{IEEE Transactions on Robotics}, vol.~36, no.~3, pp. 629--643, Jun. 2020.

\bibitem{kryczka_online_2015}
P.~Kryczka, P.~Kormushev, N.~G. Tsagarakis, and D.~G. Caldwell, ``\BIBforeignlanguage{en}{Online regeneration of bipedal walking gait pattern optimizing footstep placement and timing},'' in \emph{\BIBforeignlanguage{en}{2015 {IEEE}/{RSJ} {International} {Conference} on {Intelligent} {Robots} and {Systems} ({IROS})}}.\hskip 1em plus 0.5em minus 0.4em\relax Hamburg, Germany: IEEE, Sep. 2015, pp. 3352--3357.

\bibitem{dai2024multi}
M.~Dai, J.~Lee, and A.~D. Ames, ``Multi-domain walking with reduced-order models of locomotion,'' in \emph{2024 American Control Conference (ACC)}.\hskip 1em plus 0.5em minus 0.4em\relax IEEE, 2024, pp. 2830--2837.

\bibitem{ames2011human}
A.~D. Ames, R.~Vasudevan, and R.~Bajcsy, ``Human-data based cost of bipedal robotic walking,'' in \emph{Proceedings of the 14th international conference on Hybrid systems: computation and control}, 2011, pp. 153--162.

\bibitem{kim_once-per-step_2017}
M.~Kim and S.~H. Collins, ``Once-{Per}-{Step} {Control} of {Ankle} {Push}-{Off} {Work} {Improves} {Balance} in a {Three}-{Dimensional} {Simulation} of {Bipedal} {Walking},'' \emph{IEEE Transactions on Robotics}, vol.~33, no.~2, pp. 406--418, Apr. 2017.

\bibitem{sellaouti_faster_2006}
R.~Sellaouti, O.~Stasse, S.~Kajita, K.~Yokoi, and A.~Kheddar, ``Faster and {Smoother} {Walking} of {Humanoid} {HRP}-2 with {Passive} {Toe} {Joints},'' in \emph{2006 {IEEE}/{RSJ} {International} {Conference} on {Intelligent} {Robots} and {Systems}}, Oct. 2006, pp. 4909--4914, iSSN: 2153-0866.

\bibitem{orin_centroidal_2013}
D.~E. Orin, A.~Goswami, and S.-H. Lee, ``\BIBforeignlanguage{en}{Centroidal dynamics of a humanoid robot},'' \emph{\BIBforeignlanguage{en}{Autonomous Robots}}, vol.~35, no. 2-3, pp. 161--176, Oct. 2013.

\bibitem{Andersson2019}
J.~A.~E. Andersson, J.~Gillis, G.~Horn, J.~B. Rawlings, and M.~Diehl, ``{CasADi} -- {A} software framework for nonlinear optimization and optimal control,'' \emph{Mathematical Programming Computation}, vol.~11, no.~1, pp. 1--36, 2019.

\bibitem{wachter2006implementation}
A.~W{\"a}chter and L.~T. Biegler, ``On the implementation of an interior-point filter line-search algorithm for large-scale nonlinear programming,'' \emph{Mathematical programming}, vol. 106, pp. 25--57, 2006.

\bibitem{cassie}
Agility Robotics. \url{https://www.agilityrobotics.com/robots\#cassie}.

\bibitem{bouyarmane_quadratic_2019}
K.~Bouyarmane, K.~Chappellet, J.~Vaillant, and A.~Kheddar, ``Quadratic {Programming} for {Multirobot} and {Task}-{Space} {Force} {Control},'' \emph{IEEE Transactions on Robotics}, vol.~35, no.~1, pp. 64--77, Feb. 2019.

\bibitem{cassie_mujocosim}
Agility Robotics. \url{https://github.com/osudrl/cassie-mujoco-sim}, 2018.

\bibitem{todorov_mujoco_2012}
E.~Todorov, T.~Erez, and Y.~Tassa, ``{MuJoCo}: {A} physics engine for model-based control,'' in \emph{2012 {IEEE}/{RSJ} {International} {Conference} on {Intelligent} {Robots} and {Systems}}, Oct. 2012, pp. 5026--5033.

\bibitem{zaytsev_two_2015}
P.~Zaytsev, S.~J. Hasaneini, and A.~Ruina, ``\BIBforeignlanguage{en}{Two steps is enough: {No} need to plan far ahead for walking balance},'' in \emph{\BIBforeignlanguage{en}{2015 {IEEE} {International} {Conference} on {Robotics} and {Automation} ({ICRA})}}.\hskip 1em plus 0.5em minus 0.4em\relax Seattle, WA, USA: IEEE, May 2015, pp. 6295--6300.

\bibitem{birch_terminology_2015}
I.~Birch, W.~Vernon, J.~Walker, and M.~Young, ``Terminology and forensic gait analysis,'' \emph{Science \& Justice}, vol.~55, no.~4, pp. 279--284, Jul. 2015.

\bibitem{kojio_footstep_2020}
Y.~Kojio, Y.~Omori, K.~Kojima, F.~Sugai, Y.~Kakiuchi, K.~Okada, and M.~Inaba, ``\BIBforeignlanguage{en}{Footstep {Modification} {Including} {Step} {Time} and {Angular} {Momentum} {Under} {Disturbances} on {Sparse} {Footholds}},'' \emph{\BIBforeignlanguage{en}{IEEE Robotics and Automation Letters}}, vol.~5, no.~3, pp. 4907--4914, Jul. 2020.

\end{thebibliography}

\end{document}